\documentclass[10pt, a4paper, twocolumn]{article} 
 
\usepackage[utf8]{inputenc}
\usepackage[T1]{fontenc}
\usepackage[english]{babel}
 
\usepackage{tgheros} 

\usepackage[varqu]{zi4}
\usepackage{microtype}
 
\usepackage{titlesec}
\titlespacing*{\section}{0pt}{1.5ex plus 1ex minus .2ex}{0.1em}
\titlespacing*{\subsection}{0pt}{1.5ex plus 1ex minus .2ex}{0.1em}
\titlespacing*{\subsubsection}{0pt}{1.5ex plus 1ex minus .2ex}{0.1em}
 
\usepackage[table]{xcolor}
\usepackage{booktabs}
\usepackage{float}
\usepackage{graphicx}
\usepackage{subcaption}
\usepackage[labelfont={bf,small}, textfont={small}]{caption}
 
\definecolor{acqua_celeste}{RGB}{0, 160, 175}
 
\usepackage{amsmath}
\usepackage{amssymb}
\usepackage{mathtools}
\usepackage{amsthm}
\usepackage{bbold}
\usepackage{pifont}
\usepackage{multirow}
\usepackage{arydshln}
\usepackage{array}
\usepackage{makecell}
\usepackage{wrapfig}
\usepackage{tabularx}
 
\definecolor{pgreen}{rgb}{0.13, 0.55, 0.13}
\definecolor{pred}{rgb}{0.8, 0.13, 0.13}
\definecolor{diversity}{HTML}{D4E1F5}
\definecolor{recognizability}{HTML}{FFE6CC}
 
\newcolumntype{Y}{>{\centering\arraybackslash}X}
 
\theoremstyle{plain}

\theoremstyle{definition}

\theoremstyle{remark}

\usepackage[left=1.5cm, right=1.5cm, top=1.5cm, bottom=2cm]{geometry}
\usepackage[numbers, sort&compress]{natbib}
\usepackage{hyperref}
\usepackage[capitalize,noabbrev]{cleveref}
 
\hypersetup{
    colorlinks=true,
    linkcolor=black,
    urlcolor=acqua_celeste,
    citecolor=acqua_celeste
}

\usepackage{fancyhdr}
\usepackage{authblk}
 
\title{\huge\bfseries\vspace{-1em} Evidence-Based Text-Conditioned 3D CT Synthesis for Ovarian Cancer}
 
\author[1]{Francesca Pia Panaccione\textsuperscript{*}}
\author[1]{Eugenio Lomurno}
\author[2,3]{Francesca Fati}
\author[5]{Carlotta Pecchiari}
\author[3]{Marina Rosanu}
\author[3]{Luigi De Vitis}
\author[3]{Lucia Ribero}
\author[3]{Gabriella Schivardi}
\author[3]{Giovanni Damiano Aletti}
\author[3]{Nicoletta Colombo}
\author[4]{Maria Francesca Spadea}
\author[3]{Francesco Multinu}
\author[1]{Matteo Matteucci}
\author[2]{Elena De Momi}
 
\affil[1]{AIRLab, Politecnico di Milano, Milan, Italy}
\affil[2]{NEARLab, Politecnico di Milano, Milan, Italy}
\affil[3]{Istituto Europeo di Oncologia, Milan, Italy}
\affil[4]{Karlsruhe Institute of Technology, Karlsruhe, Germany}
\affil[5]{Universit\`a degli Studi dell'Insubria, Varese, Italy}
 
\date{}
 
\begin{document}
 
\twocolumn[
  \begin{@twocolumnfalse}
    \maketitle
    \vspace{-2em}
 
    \begin{abstract}
        \setlength{\parindent}{0pt}
        \setlength{\parskip}{4pt}
        \itshape
        \noindent Ovarian cancer is frequently diagnosed at an advanced stage, making preoperative contrast-enhanced computed tomography (CT) central to disease staging and surgical planning; yet the scarcity of annotated imaging data, compounded by privacy regulations, limits the development of generalizable computational models in this domain. Text-conditioned three-dimensional CT synthesis has recently shown promise for synthetic data generation, but existing pipelines depend on paired radiology reports and have been evaluated exclusively on chest CT. We propose OvESyn (Ovarian Evidence-based Synthesis), a framework that constructs standardized Findings and Impression sections directly from CT-derived imaging descriptors and routinely available clinical metadata, without accessing any original radiology report, and uses these structured descriptions to condition a latent diffusion model adapted to a cohort of 493 high-grade serous ovarian carcinoma patients. To our knowledge, this is the first text-conditioned 3D CT synthesis framework adapted to an abdomino-pelvic oncologic setting. A systematic ablation over the two adaptation axes---vision-language encoder alignment and generator fine-tuning---identifies generator domain adaptation as the operative mechanism for crossing the domain gap and establishing the target anatomical structure: without it, synthesis remains anchored to the thoracic pretraining domain, with Precision and Recall collapsing to zero and FID2.5D exceeding 140, regardless of encoder alignment. Encoder alignment, which does not by itself establish anatomical structure, instead refines intensity and fine detail. The full OvESyn model attains the best distributional and intensity fidelity (FID2.5D 29.35, Precision 0.671, Wasserstein-1 0.044), while the variant adapting only the generator maximizes coverage (Recall 0.645), reflecting a fidelity--coverage trade-off governed by encoder adaptation. Requiring no paired radiology reports, only automatic segmentations and routine preoperative metadata, OvESyn supports transferability to report-scarce institutional setting and provides a foundation for synthetic cohort generation in abdomino-pelvic oncologic imaging.
    \end{abstract}
 
    \vspace{0.5em}
    \noindent\textbf{Keywords:} computed tomography synthesis $\cdot$ data augmentation $\cdot$ domain adaptation $\cdot$ generative models $\cdot$ high-grade serous ovarian carcinoma $\cdot$ latent diffusion $\cdot$ report generation $\cdot$ vision-language alignment
 
    \vspace{1em}
    \hrule height 1pt
    \vspace{2em}
  \end{@twocolumnfalse}
]
 
{
  \renewcommand{\thefootnote}{\fnsymbol{footnote}}
  \footnotetext[1]{Corresponding author: \texttt{francesca.panaccione@polimi.it}. Code available at \url{https://github.com/francescapia/OvESyn}.}
}
 
\section{Introduction}
\label{sec:introduction}
Ovarian cancer ranks eighth in cancer incidence and mortality among women worldwide \cite{bray2024global}, with approximately 70\% of cases diagnosed at an advanced stage due to the absence of specific early symptoms \cite{hong2025early,bergin2024time}. Preoperative contrast-enhanced computed tomography (CT) of the thorax, abdomen, and pelvis is the primary imaging modality for disease staging and treatment planning according to current clinical guidelines \cite{rizzo2025ovarian,ledermann2024esgo}. Staging relies on the International Federation of Gynecology and Obstetrics (FIGO) classification, where the distinction between stage III and stage IV disease, determined by CT-detectable extraperitoneal spread, informs the choice between primary debulking surgery and neoadjuvant chemotherapy \cite{miceli2023imaging,tsili2024imaging}.
 
Institutional datasets in this setting are inherently small: acquisition requires dedicated preoperative protocols, annotation demands specialized radiological expertise, and data sharing is constrained by privacy regulations including the General Data Protection Regulation \cite{hansen2021assessment,bajada2025gdpr,narayan2025addressing}. These factors collectively limit the development of data-hungry deep learning models \cite{lecun2015deep} and restrict the transferability of trained systems across institutions. Synthetic data generation has been proposed to address both constraints, supporting data augmentation and privacy-preserving cohort sharing \cite{dumont2021overcoming,jiang2025synthetic,giouroukou2025rethinking}; among these approaches, text-conditioned 3D CT synthesis is particularly attractive because it grants semantic control over the generated volumes through natural language.
 
Recent latent diffusion models conditioned on radiology reports produce anatomically coherent chest CT volumes \cite{hamamci2024generatect,xu2024medsyn,molino2025text}, yet they share two limitations for the present setting: they have been developed and evaluated exclusively on chest CT, and they depend structurally on paired radiology reports for conditioning. Transferring this paradigm to abdomino-pelvic oncology is far from straightforward: datasets are an order of magnitude smaller, and the anatomical complexity of multi-site abdomino-pelvic disease substantially exceeds that of the thorax. The decisive obstacle, however, is the conditioning text itself, which in this domain is heterogeneous, encodes institution-specific conventions, and is frequently unavailable in a standardized form.
 
We address this barrier by constructing the conditioning text directly from imaging evidence rather than retrieving it from a report. We introduce OvESyn (Ovarian Evidence-based Synthesis), a framework that extracts CT-based imaging descriptors (intensity statistics, morphological features, and spatial context obtained from automatic tumor and organ segmentations) and, combined with two routinely collected clinical variables, serializes them into standardized Findings and Impression sections through a large language model \cite{yang2025qwen3}; these structured sections then condition a latent diffusion model \cite{molino2025text} adapted to a cohort of 493 high-grade serous ovarian carcinoma (HGSOC) patients. The two clinical variables, FIGO stage and ascites status, are intentionally kept minimal: routinely available, unambiguous, and consistent across institutions \cite{szender2017impact,andreou2023prognostic}.
 
Beyond the conditioning interface, adapting the framework to a new, data-scarce domain involves two distinct axes: vision-language alignment of the encoder, and adaptation of the generative model to the target anatomy. Our experiments show these axes to be asymmetric and complementary, governing different aspects of the output. Full fine-tuning of the generator is the operative mechanism for crossing the domain gap: it migrates synthesis from the thoracic pretraining domain to the abdomino-pelvic target and establishes the anatomical (macro) structure, producing anatomically coherent volumes even from a small cohort; without it, generation remains anchored to the pretraining domain regardless of encoder alignment. Vision-language alignment does not establish anatomical structure on its own; it acts as a complementary refinement of intensity and fine detail, improving per-sample realism (Precision) at the cost of coverage (Recall). The two are therefore best combined: the generator supplies the anatomical structure, the encoder refines intensity and fine detail within it.
 
The contributions of this work are as follows:
\begin{itemize}
\item A structured report generation pipeline that constructs standardized Findings and Impression sections from CT volume-derived descriptors and routine clinical metadata, removing the dependency on paired radiology reports.
\item OvESyn, the first systematic adaptation of a text-conditioned 3D CT generation framework to an abdomino-pelvic oncologic setting, demonstrated on a cohort of 493 HGSOC patients.
\item A controlled ablation characterizing the functional specialization between the two adaptation axes in a data-scarce transfer setting: full generator fine-tuning is the crucial adaptation for crossing the domain gap and establishing anatomical structure, whereas vision-language alignment provides a complementary refinement of intensity and fine detail, a finding with implications for text-conditioned generative model design beyond ovarian CT.
\end{itemize}
 
\section{Related Work}
\label{sec:related_work}
 
\subsection{Text-Conditioned 3D CT Generation}
Text-conditioned approaches represent a growing but still minority paradigm in knowledge-guided 3D CT generation, where geometric mask conditioning has dominated \cite{panaccioneknowledge}. Linguistic descriptions offer a qualitatively different interface: they encode semantic content, disease extent, and clinical context without requiring voxel-aligned spatial priors, and their construction is increasingly tractable owing to advances in pretrained language models. GenerateCT \cite{hamamci2024generatect} demonstrated that free-form radiology text can condition cascaded latent diffusion to produce coherent chest CT volumes at scale, using the CT-RATE dataset of over 25{,}000 paired scan-report entries. MedSyn \cite{xu2024medsyn} combined textual and anatomical layout conditioning in a multi-stage architecture for high-fidelity lung synthesis. More recently, single-stage latent diffusion with contrastive 3D vision-language pretraining has been shown to produce anatomically consistent volumes from text alone \cite{molino2025text}. All these methods share a structural dependency on paired radiology reports for conditioning and have been developed and evaluated exclusively on thoracic imaging. Their transfer to abdomino-pelvic oncology is precluded by domain shift and by the scarcity of report-paired datasets in this setting.
 
\subsection{Structured Report Generation from Medical Imaging}
Structured reporting has been established as a standard for reducing variability and improving clinical communication in radiology \cite{european2023esr}, and vision-language pretraining methods have demonstrated that aligned image-text representations support downstream report generation tasks \cite{boecking2022making,wang2022medclip}. These works treat report generation as an end product of radiological interpretation. In OvESyn, structured text is instead derived algorithmically from imaging evidence as an integrated conditioning interface within the generative pipeline --- not to produce a paired volume-report dataset, but to synthesize a semantically grounded signal that drives volumetric generation directly. No prior work has proposed this formulation.
 
\subsection{AI in Ovarian Cancer CT}
Computational methods applied to ovarian cancer CT have concentrated on discriminative tasks, including tumor segmentation \cite{buddenkotte2023deep}, multi-organ anatomical parsing \cite{wasserthal2023totalsegmentator}, treatment response prediction \cite{huang2024predicting}, cytoreduction outcome estimation \cite{egger2022predicting}, and diagnostic characterization of ovarian masses \cite{adusumilli2025methodological}. No generative approach has been applied to ovarian cancer CT, leaving the design space for volumetric synthesis in abdomino-pelvic oncology entirely unexplored.
 
\begin{figure*}[t]
    \centering
    \includegraphics[width=\textwidth]{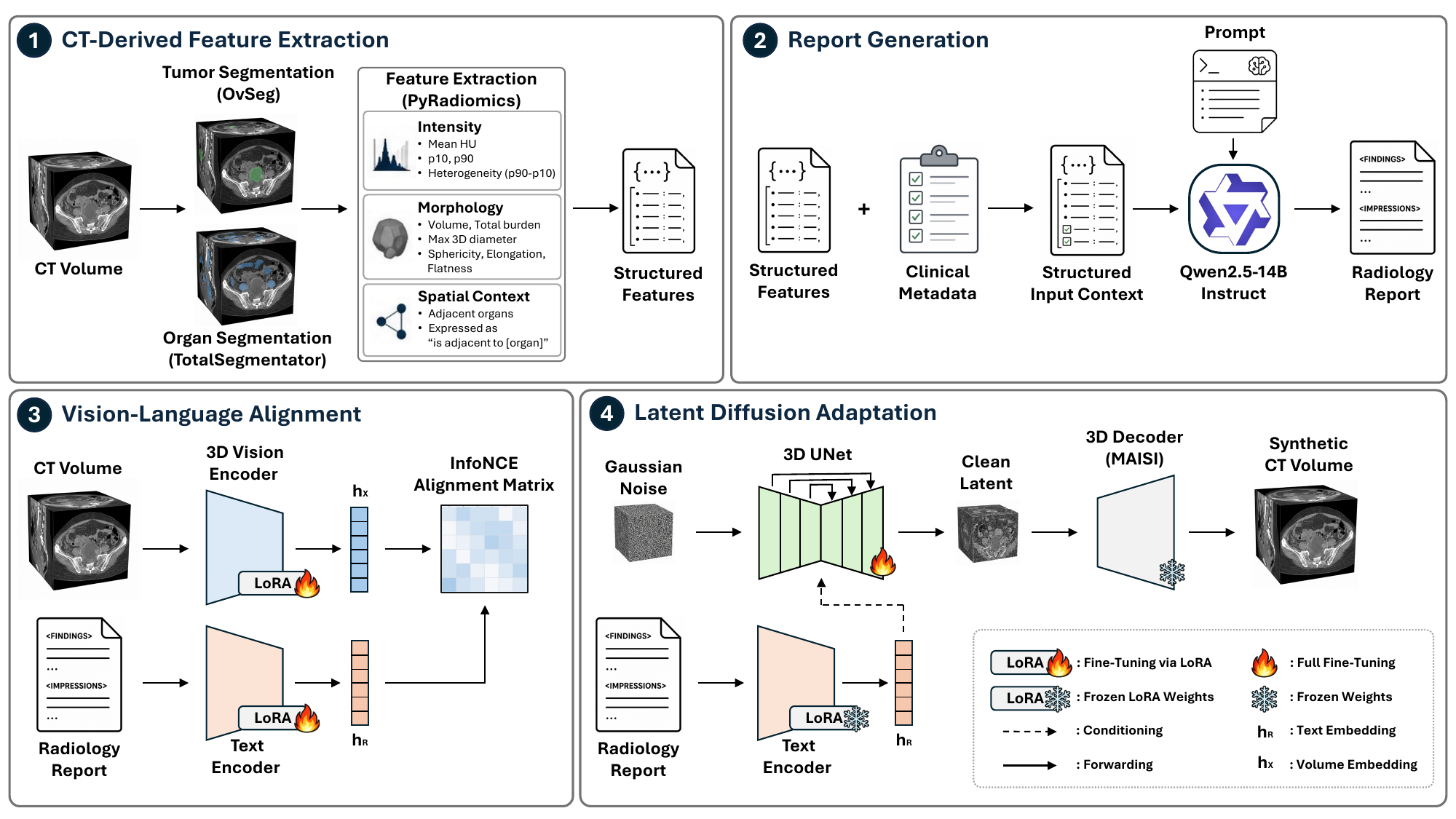}
    \caption{OvESyn pipeline overview. The structured report generation stage (left, primary contribution) extracts CT-derived features from automatic segmentations and integrates routine clinical metadata to construct standardized Findings and Impression text. The text-conditioned generation stage (right) encodes the report via a 3D-CLIP encoder and synthesizes a volumetric CT output through a latent diffusion model.}
    \label{fig:pipeline}
\end{figure*}
 
\section{Method}
\label{sec:method}
 
\textbf{Overview.} Let $\mathbf{x} \in \mathbb{R}^{H \times W \times D}$ denote a 3D CT volume and $\mathbf{m} = \{\text{FIGO stage},\ \text{ascites status}\}$ a set of routine clinical metadata collected during standard preoperative workup. Given $\mathbf{x}$, a tumor segmentation $\mathbf{s}_t$ obtained via OvSeg \cite{buddenkotte2023deep}, and an organ segmentation $\mathbf{s}_o$ obtained via TotalSegmentator \cite{wasserthal2023totalsegmentator}, OvESyn constructs a structured radiology report:
\begin{equation}
    r = \texttt{ReportGen}(\mathbf{x},\ \mathbf{s}_t,\ \mathbf{s}_o,\ \mathbf{m}),
\end{equation}
which a text encoder $f_\mathcal{R}$ maps to a conditioning embedding $f_\mathcal{R}(r)$. This embedding conditions a generative model $p_\theta$, parameterized by $\theta$, that defines a probability distribution over CT volumes, from which a synthetic volume $\hat{\mathbf{x}}$ is sampled:
\begin{equation}
    \hat{\mathbf{x}} \sim p_\theta\!\left(\mathbf{x} \mid f_\mathcal{R}(r)\right).
\end{equation}
The synthesized volume $\hat{\mathbf{x}}$ is not a reconstruction of an existing scan but a new sample whose anatomical and pathological content is governed by the descriptors encoded in $r$. We instantiate $p_\theta$ as a latent diffusion model: conditioned on $f_\mathcal{R}(r)$, it iteratively denoises a randomly initialized latent code, which a decoder maps back to voxel space, so that repeated sampling under the same report yields distinct yet clinically consistent volumes. The structured report $r$ is thus the interface through which imaging and clinical evidence steer the generative process. The framework comprises two sequential stages (Fig.~\ref{fig:pipeline}): a structured report generation stage, which produces $r$ from CT-derived features and routine clinical metadata, and a text-conditioned generation stage, which adapts an existing latent diffusion framework \cite{molino2025text} to the ovarian cancer domain.
 
\textbf{CT-Derived Feature Extraction.} All imaging descriptors are computed deterministically from the CT volume and its lesion and organ segmentation masks using PyRadiomics \cite{van2017computational}. \textit{Intensity descriptors} are measured within each tumor mask in Hounsfield units: mean intensity, 10th and 90th percentiles, and heterogeneity spread (p90$-$p10)~\cite{prokop2003spiral}. These statistics are mapped to qualitative density labels (predominantly solid, mixed, or complex cystic based on mean intensity~\cite{lupean2020computer}; mildly, moderately, or markedly heterogeneous based on spread~\cite{park2020reliability,vargas2017novel}) to ensure vocabulary consistency across patients. \textit{Morphological descriptors} are measured directly from each lesion's segmentation mask: lesion volume (mL), total tumor burden as the sum of all lesion volumes, maximum 3D diameter (mm), and shape indices (sphericity, elongation, and flatness) mapped to categorical descriptors (spherical/ovoid, elongated, flattened, lobulated)~\cite{van2017computational,leng2024development}. \textit{Spatial context} is computed from the lesion and organ segmentations: for each lesion, the set of anatomically adjacent organs is identified and expressed using constrained proximity phrasing, such as \textit{``is adjacent to [organ]''} or \textit{``adjacent to [organ]''}, preventing unsupported claims of invasion and reducing inter-patient inconsistency.
 
\textbf{Report Generation.} Two routine clinical variables are provided alongside the imaging descriptors: FIGO stage, which encodes disease extent and correlates with peritoneal tumor distribution patterns visible on CT \cite{miceli2023imaging}, and ascites status, a marker of disease burden with established prognostic significance \cite{szender2017impact}. Both are recorded in the standard preoperative workup without additional annotation overhead and are available across institutions independently of local reporting conventions. The CT-derived descriptors and these two variables are serialized into a structured input context and submitted to Qwen2.5-14B-Instruct \cite{yang2025qwen3} through separate prompt templates for the Findings and Impression sections. FIGO stage appears exclusively as a correlate statement; ascites is expressed with fixed wording. The model output is parsed via structured tags \texttt{<FINDINGS>} and \texttt{<IMPRESSION>}, yielding a report in standard radiological style. Table~\ref{tab:report_example} shows a representative generated report for a Stage~IV patient with bilateral lesions and ascites. Full prompt templates are provided in Appendix~A.
 
\begin{table*}[t]
\caption{Representative structured report generated by OvESyn for a Stage~IV HGSOC patient. No original radiology report was accessed.}
\label{tab:report_example}
\centering
\begin{tabular}{p{0.16\textwidth} p{0.78\textwidth}}
\toprule
\textbf{Section} & \textbf{Generated Text} \\
\midrule
Findings & An irregular multilobulated omental mass measuring 28~cm with volume 177.3~mL demonstrates mixed solid and cystic attenuation with moderate heterogeneity. The mass abuts small bowel and colon. A separate lobulated pelvic/ovarian mass measuring 12.6~cm with volume 274.8~mL shows mixed solid and cystic components with marked heterogeneity, including low attenuation areas, abutting small bowel, colon, and urinary bladder. Combined tumor burden is 452.1~mL. Ascites is present. Findings are consistent with FIGO Stage~IV disease. \\
\midrule
Impression & Extensive multifocal omental and adnexal/pelvic tumor burden totaling 452.1 mL, with ascites. Markedly heterogeneous mixed solid-cystic appearance suggests FIGO stage IV disease. \\
\bottomrule
\end{tabular}
\end{table*}
 
\textbf{Vision-Language Alignment.} Text-conditioned generation is built on a 3D-CLIP architecture \cite{molino2025text} comprising a 3D vision encoder and a masked self-attention text encoder, jointly trained to align CT volumes and radiology reports in a shared embedding space. Alignment is optimized via a symmetric InfoNCE objective \cite{oord2018representation}:
\begin{equation}
    \mathcal{L}_{\text{CLIP}} = \mathcal{L}_{\mathcal{X} \to \mathcal{R}} + \mathcal{L}_{\mathcal{R} \to \mathcal{X}},
    \label{eq:clip}
\end{equation}
where $\mathcal{L}_{\mathcal{X} \to \mathcal{R}}$ and $\mathcal{L}_{\mathcal{R} \to \mathcal{X}}$ denote the image-to-text and text-to-image contrastive terms respectively, computed over L2-normalized embeddings with a learnable temperature parameter. Both encoders are domain-adapted via Low-Rank Adaptation (LoRA) \cite{hu2022lora} on the query and value attention projections: restricting updates to low-rank perturbations preserves the representational structure learned during chest-CT pretraining while enabling parameter-efficient fine-tuning on the limited HGSOC cohort.
 
\textbf{Latent Diffusion Adaptation.} The generative backbone follows a latent diffusion architecture \cite{rombach2022high}: a MAISI variational autoencoder \cite{guo2025maisi} compresses CT volumes into a compact latent representation, and a 3D U-Net \cite{molino2025text} denoises latent codes conditioned on the text embedding $\mathbf{h}_\mathcal{R} = f_\mathcal{R}(r)$ via cross-attention at multiple resolutions. At inference, classifier-free guidance \cite{ho2022classifier} balances generation fidelity and diversity. In contrast to the LoRA-adapted encoders, the U-Net is fully fine-tuned from its pretrained checkpoint, since it must learn the abdomino-pelvic volumetric distribution itself, whereas the encoders require only a lightweight domain adaptation of their pretrained representations.
 
\section{Experimental Setup}
\label{sec:experimental_setup}
This section describes the dataset, the ablation configurations evaluated to characterize OvESyn's adaptation components, and the evaluation protocol.
 
\subsection{Dataset}
\textbf{Cohort and acquisition.} The study cohort comprises 493 patients with histologically confirmed HGSOC, retrospectively collected at the Istituto Europeo di Oncologia (IEO), Milan. All patients underwent preoperative contrast-enhanced CT of the thorax, abdomen, and pelvis in the portal venous phase. The study was approved by the Scientific Board of IEO under protocol UID 4134. All patients provided informed consent for the use of their data for research purposes. The dataset is split into 344 training, 73 validation, and 76 test cases, stratified by acquisition year to prevent temporal leakage. Tumor segmentations are produced by OvSeg \cite{buddenkotte2023deep}; organ segmentations are obtained via TotalSegmentator \cite{wasserthal2023totalsegmentator}.
 
\textbf{Preprocessing.} All volumes follow a fixed protocol: abdomino-pelvic region of interest cropping from the ischiopubic rami to the hemidiaphragm, soft-tissue windowing (level 40~HU, width 400~HU), background correction, denoising, intensity normalization to $[0,1]$, isotropic resampling to $1{\times}1{\times}1$~mm, and reshape to $512{\times}512{\times}128$ voxels.
 
\subsection{Training Configurations}
The ablation study varies two independent adaptation axes in OvESyn: CLIP encoder fine-tuning and U-Net generator fine-tuning. Its purpose is to disentangle how each of these two asymmetric adaptations --- LoRA on the 3D-CLIP encoder versus full fine-tuning of the U-Net generator --- contributes to synthesis quality in this low-data regime. The full model, OvESyn, applies LoRA adaptation to the 3D-CLIP encoder and full fine-tuning to the U-Net generator. Three ablated variants isolate the contribution of each component: OvESyn$_C$ adapts only the CLIP encoder (LoRA), keeping the U-Net at its pretrained chest-CT weights; OvESyn$_U$ adapts only the U-Net generator (full fine-tuning), keeping the CLIP encoder frozen; and OvESyn$_\emptyset$ applies no domain adaptation to either component, serving as the lower bound. All variants share the same data split, preprocessing, report generation pipeline, and inference protocol. Training hyperparameters are reported in Appendix~B; all experiments are conducted on a single GPU node equipped with 8 $\times$ NVIDIA A100 GPUs, each with 40~GB of HBM2 memory. As an external point of comparison, MedSyn \cite{xu2024medsyn} is adapted to the same cohort and preprocessing protocol via full fine-tuning of its BERT text encoder and low-resolution generation stage, leaving the super-resolution module unchanged. Given its qualitatively different multi-stage architecture, it serves as an external reference rather than a direct ablation baseline.
 
\subsection{Evaluation Metrics}
\textbf{Generation metrics.} Feature-space realism is assessed by three complementary measures: \textit{FID2.5D}, computed from InceptionV3 \cite{szegedy2016rethinking} features over 24 slices per anatomical axis, averaged across axes \cite{heusel2017gans}; \textit{FID3D}, computed as the Fr\'echet distance between per-channel statistics of VAE latent representations \cite{ellis2022evaluation}, providing a domain-specific volumetric complement; and \textit{Improved Precision and Recall} \cite{kynkaanniemi2019improved} ($k{=}5$), decomposing distributional similarity into per-sample fidelity (Precision) and coverage (Recall). Intensity fidelity is evaluated via the \textit{Wasserstein-1 distance} (W1; 1001-bin histogram over $[0,1]$) and the \textit{mean intensity error} $\Delta\mu$, the absolute difference in mean voxel intensity between generated and real volumes.
 
\section{Results}
\label{sec:results}
Results follow the pipeline structure: report generation quality, vision-language alignment, generation quality, and the functional specialization of the two adaptation axes.
 
\subsection{Report Generation Quality}
\begin{figure*}[!t]
    \centering
    \includegraphics[width=0.99\linewidth]{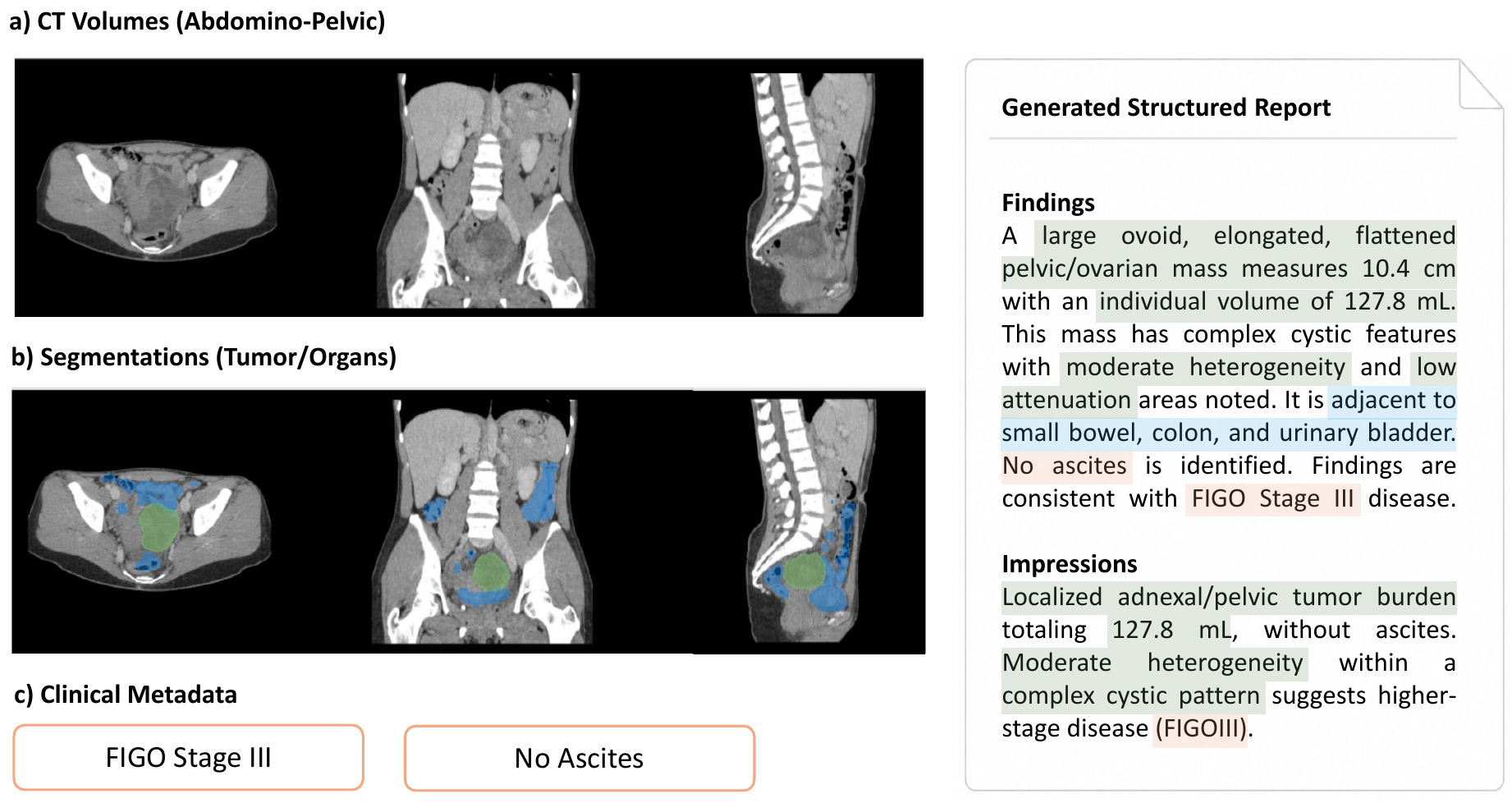}
    \caption{Structured report generation for a representative Stage~III HGSOC patient without ascites. Segmentation overlays link CT-derived imaging evidence to the generated Findings and Impression, including tumor morphology, organ contact, ascites status, and FIGO stage. No original radiology report was used.}
    \label{fig:report_example}
\end{figure*}

The structured report generation pipeline produces clinically consistent Findings and Impression sections without accessing any original radiology report. Because the descriptors are grounded in imaging evidence and expressed through a constrained vocabulary, the resulting reports are specific to individual tumor morphology, spatial distribution, and disease burden. Fig.~\ref{fig:report_example} illustrates a representative Stage~III case with a single ovarian/pelvic mass and no ascites: colored overlays directly link each segmented region to the corresponding textual descriptor, with volumetric measurements, density labels, and organ abutments in the generated text traceable to specific imaging findings. Appendix~C extends this analysis to cases spanning FIGO stage III and IV, varying ascites status, and single versus bilateral lesion configurations, confirming that this consistency is not specific to the case shown.
 
\subsection{Vision-Language Alignment}
CLIP encoder fine-tuning produces a marked improvement in alignment (Table~\ref{tab:clip}). CLIPScore rises from 0.263 to 0.993 and the contrastive loss drops from 3.551 to 1.058, reflecting near-complete alignment between paired CT and text embeddings. Text-to-image retrieval, however --- the relevant direction for a text-conditioned generator --- remains close to chance even after adaptation: Recall@1 reaches only 0.026, about twice the $1/76$ random level for $N{=}76$, while Recall@5 and Recall@10 stay at 0.092 and 0.184, roughly $1.4\times$ chance. This co-occurrence of near-perfect pairwise alignment with near-random retrieval reflects the cohort's homogeneity: clinical prompts share highly overlapping disease descriptors by construction, so alignment can concentrate each volume onto its own report without rendering distinct cases separable. Linear probing on the frozen image embeddings improves from 0.442 to 0.561 AUC for FIGO stage and from 0.483 to 0.607 AUC for ascites status; the larger, clearly above-chance gain for ascites is consistent with its diffuse, high-contrast imaging signature, whereas the subtler stage III/IV boundary --- which hinges on small extraperitoneal sites --- is harder to encode in a global volume embedding. Notably, the pre-adaptation FIGO AUC lies below chance, suggesting that LoRA does not merely add discriminative signal but corrects a representation misaligned with the abdomino-pelvic domain.
 
\begin{table*}[!t]
\caption{CLIP evaluation on the test split ($N{=}76$), comparing the chest-CT pretrained encoder (Pretrained) with its LoRA-adapted counterpart (LoRA-adapted). CLIP Loss is the contrastive alignment objective and CLIP Score the cosine similarity between paired image and text embeddings; R@$K$ reports text-to-image retrieval. FIGO and ascites AUC are obtained by linear probing of the frozen image embeddings. $\uparrow$: higher is better; $\downarrow$: lower is better. \textbf{Bold}: best value per metric.}
\label{tab:clip}
\centering
\small
\renewcommand{\arraystretch}{1.2}
\begin{tabular*}{\textwidth}{@{\extracolsep{\fill}}lccccccc@{}}
\toprule
\textbf{Model} &
\textbf{Loss $\downarrow$} &
\textbf{CLIP $\uparrow$} &
\textbf{R@1 $\uparrow$} &
\textbf{R@5 $\uparrow$} &
\textbf{R@10 $\uparrow$} &
\textbf{FIGO AUC $\uparrow$} &
\textbf{Asc. AUC $\uparrow$} \\
\midrule
Pretrained   & 3.5513 & 0.2628 & 0.0132 & 0.0658 & 0.1184 & 0.4416 & 0.4833 \\
LoRA-adapted & \textbf{1.0576} & \textbf{0.9925} & \textbf{0.0263} & \textbf{0.0921} & \textbf{0.1842} & \textbf{0.5613} & \textbf{0.6065} \\
\bottomrule
\end{tabular*}
\end{table*}
 
\subsection{Generation Quality}
Generator fine-tuning is the decisive factor for synthesis quality in this regime (Table~\ref{tab:generation}). The two configurations without generator adaptation, OvESyn$_\emptyset$ and OvESyn$_C$, fail to cross the domain gap: both yield FID2.5D above 140 with Precision and Recall at zero, and qualitatively both reproduce thoracic anatomy --- recognizable lung fields rather than the abdomino-pelvic target --- regardless of encoder alignment (Fig.~\ref{fig:qualitative}, columns a--b). CLIP fine-tuning alone (OvESyn$_C$) lowers intensity error relative to OvESyn$_\emptyset$ (W1 $0.157\!\to\!0.094$, $\Delta\mu$ $0.124\!\to\!0.062$) and visibly sharpens the rendering, but does not alter the underlying anatomy: the output remains thoracic and, consistently across cases, laterally flipped. Once the generator is fine-tuned, both OvESyn$_U$ and the full OvESyn produce recognizable abdomino-pelvic anatomy across all three planes (Fig.~\ref{fig:qualitative}, columns c--d), with FID2.5D below 33 and FID3D dropping by more than an order of magnitude (from $0.06$--$0.08$ to $0.003$); FID3D is in fact identical ($0.0032$) for the two adapted configurations, saturating once the domain gap is crossed and providing little discrimination between them. Among these two, the full model attains the best feature-space realism (FID2.5D 29.35), per-sample fidelity (Precision 0.671), and intensity fidelity (W1 0.044, $\Delta\mu$ 0.038), while OvESyn$_U$ attains the highest coverage (Recall 0.645). MedSyn, adapted under the same protocol, underperforms every OvESyn configuration across all metrics, consistent with domain shift accumulating across its multi-stage architecture.
 
\subsection{Functional Asymmetry of the Adaptation Axes}
\begin{figure*}[!t]
    \centering
    \includegraphics[width=1\textwidth]{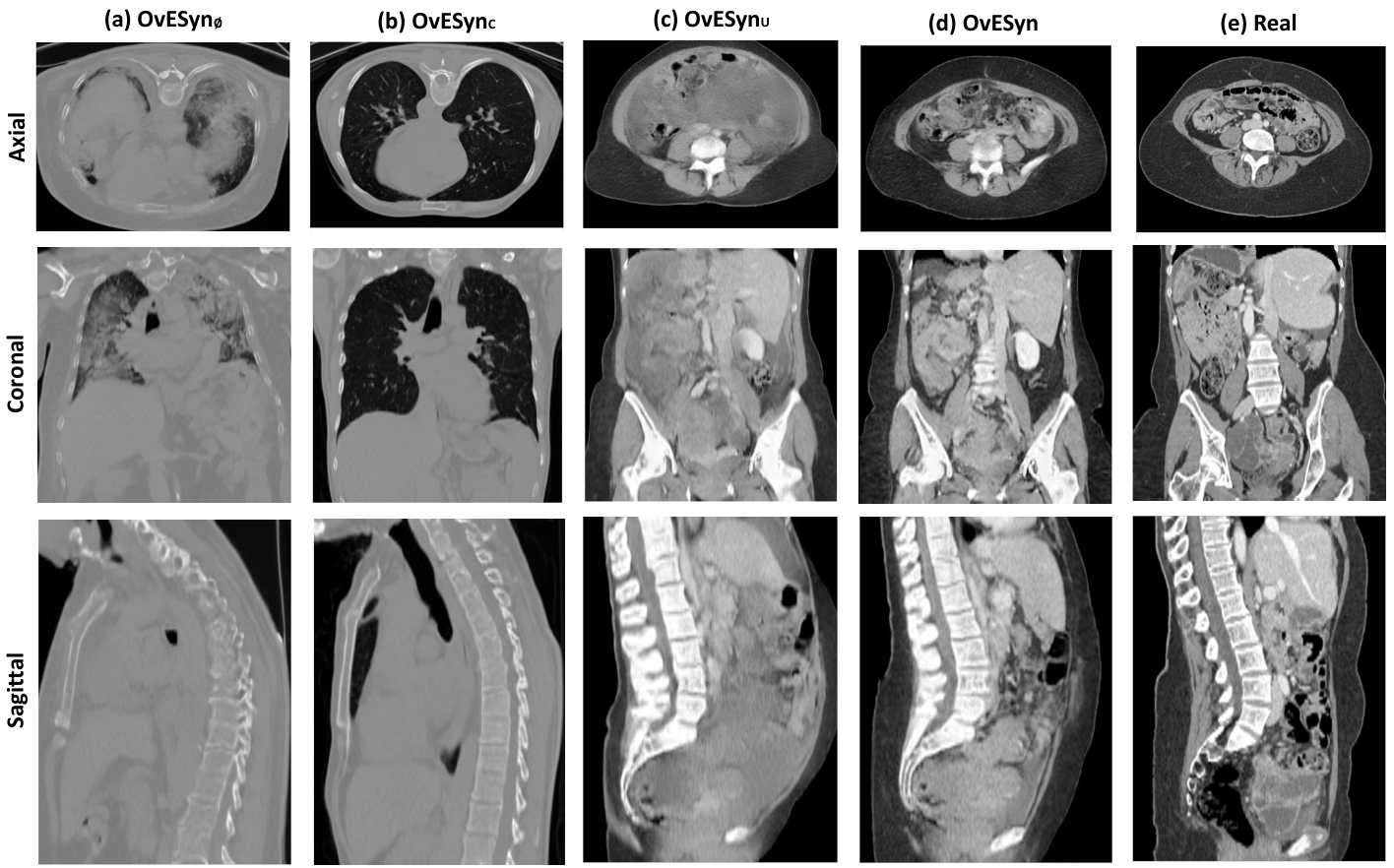}
    \caption{Qualitative comparison on a representative test case. Columns show the four OvESyn configurations and the reference CT (Real); rows report axial, coronal, and sagittal views, all displayed with identical slice indices, crop regions, and intensity window computed from the reference. Without generator fine-tuning, OvESyn$_\emptyset$ and OvESyn$_C$ generate thoracic anatomy --- recognizable lung fields rather than the abdomino-pelvic target --- and OvESyn$_C$ additionally produces consistently left--right flipped volumes; CLIP fine-tuning sharpens the rendering but does not cross the domain gap. Generator fine-tuning (OvESyn$_U$ and OvESyn) yields recognizable abdomino-pelvic anatomy across all planes, with the full OvESyn producing the most realistic volume and the closest match to the reference.}
    \label{fig:qualitative}
\end{figure*}

Considered jointly, the alignment and generation analyses (Tables~\ref{tab:clip},~\ref{tab:generation}; Fig.~\ref{fig:qualitative}) reveal a pronounced functional asymmetry between the two adaptation axes. Generator fine-tuning is the determining factor for domain transfer: it is both necessary and sufficient to relocate the synthesized distribution from the thoracic pretraining manifold to the abdomino-pelvic target, and encoder alignment cannot substitute for it --- OvESyn$_C$, despite a near-perfect CLIPScore of 0.993, continues to generate thoracic volumes with Precision and Recall identically zero. Encoder alignment, conversely, does not contribute anatomical structure; its effect is confined to the refinement of intensity and per-sample fidelity once the target anatomy has been established. This distinction is quantified by the two adapted configurations: augmenting OvESyn$_U$ with encoder alignment increases Precision from 0.434 to 0.671 and reduces both intensity errors (W1 and $\Delta\mu$), at the expense of Recall, which decreases from 0.645 to 0.421. The product of Precision and Recall is, however, essentially preserved across the two configurations (0.280 versus 0.283), indicating that encoder alignment does not expand the joint fidelity--coverage capacity of the generator but reallocates it toward per-sample fidelity. At this preserved capacity the full model further attains the lowest FID2.5D and the smallest intensity deviation, and is therefore the preferred configuration: it equals OvESyn$_U$ in aggregate fidelity--coverage while improving distributional and intensity realism. The attendant reduction in coverage is consistent with the low diversity of the conditioning reports in this cohort, a point developed in Section~\ref{sec:discussion}.
 
\begin{table*}[!t]
\caption{Generation quality on the test split ($N{=}76$). OvESyn$_\emptyset$, OvESyn$_C$, and OvESyn$_U$ denote no adaptation, CLIP-only adaptation, and U-Net-only adaptation, respectively; OvESyn applies both. FID2.5D and FID3D assess feature-space distributional realism; Precision and Recall decompose distributional similarity into per-sample fidelity and coverage~\cite{kynkaanniemi2019improved}; W1 and $\Delta\mu$ measure intensity fidelity (Wasserstein-1 distance and mean intensity error). MedSyn is an external multi-stage reference. $\uparrow$: higher is better; $\downarrow$: lower is better. \textbf{Bold}: best value per metric. The full OvESyn configuration (highlighted) is the preferred model.}
\label{tab:generation}
\centering
\small
\renewcommand{\arraystretch}{1.25}
\begin{tabular*}{\textwidth}{@{\extracolsep{\fill}}lcccccc@{}}
\toprule
\textbf{Model} &
\textbf{FID2.5D $\downarrow$} &
\textbf{FID3D $\downarrow$} &
\textbf{Prec.\ $\uparrow$} &
\textbf{Rec.\ $\uparrow$} &
\textbf{W1 $\downarrow$} &
\textbf{$\Delta\mu$ $\downarrow$} \\
\midrule
MedSyn & 243.2152          & 0.35109          & 0.0000          & 0.0000          & 0.41090          & 0.4067    \\
OvESyn$_\emptyset$ & 146.5721 & 0.0645 & 0.0000 & 0.0000 & 0.1571 & 0.1241 \\
OvESyn$_C$         & 141.8698 & 0.0813 & 0.0000 & 0.0000 & 0.0937 & 0.0618 \\
OvESyn$_U$         & 31.6891 & \textbf{0.0032} & 0.4342 & \textbf{0.6447} & 0.0450 & 0.0391 \\
\textbf{OvESyn}    & \textbf{29.3495} & \textbf{0.0032} & \textbf{0.6711} & 0.4211 & \textbf{0.0439} & \textbf{0.0375} \\
\bottomrule
\end{tabular*}
\end{table*}
 
\section{Discussion}
\label{sec:discussion}
OvESyn demonstrates that text-conditioned 3D CT synthesis for ovarian cancer is achievable without paired radiology reports, by deriving conditioning signals algorithmically from CT-derived imaging features and minimal routine clinical metadata. Three implications emerge from the results.
 
\textbf{Structured report generation as a transferable conditioning interface.} The pipeline requires only automatic segmentations --- producible with TotalSegmentator and OvSeg, both publicly available --- and two variables collected in any standard preoperative workup. This removes the institutional dependency on structured radiology reports, which are frequently absent, heterogeneous, or non-shareable across centers. Any institution with preoperative HGSOC CT data and FIGO staging records can apply OvESyn without additional annotation effort, making synthetic cohort generation feasible in data-scarce and report-scarce settings alike.
 
\textbf{Complementary roles of encoder alignment and generator adaptation.} The ablation separates two adaptation axes with distinct and complementary functions. Full fine-tuning of the generator bridges the domain gap, relocating synthesis from the thoracic pretraining manifold to the abdomino-pelvic target; encoder alignment does not contribute anatomical structure but refines intensity and per-sample fidelity within it. Because the two axes act on disjoint aspects of the output, their benefits accrue without interference: the full model combines the correct anatomy supplied by the generator with the improved intensity and precision supplied by the encoder. The accompanying reduction in coverage is, we argue, a consequence of the limited diversity of the conditioning text. Constructed from a constrained vocabulary over a histologically homogeneous HGSOC population, the generated reports are highly similar across patients, and a strongly aligned encoder maps such near-identical prompts to near-identical embeddings, thereby narrowing the diversity of the conditioned outputs. The precision--coverage trade-off between OvESyn$_U$ and the full model is, in this light, expected, and it points to the diversity of the conditioning signal --- rather than encoder alignment itself --- as the factor governing coverage. More broadly, in data-scarce transfer settings full adaptation of the generator is the dependable mechanism for crossing the domain gap, while encoder alignment is most appropriately regarded as a complementary refinement whose influence on coverage is mediated by the diversity of the conditioning signal.
 
\textbf{Scope and future directions.} OvESyn is a first step toward text-conditioned volumetric synthesis in abdomino-pelvic oncology. On the conditioning side, the limited diversity of the generated reports is the most direct constraint on coverage: enriching the descriptors --- for instance with additional routinely available clinical variables or less constrained phrasing --- could broaden the conditioned output distribution. On the vision-language alignment side, more expressive contrastive frameworks such as SigLIP could improve embedding quality under the class imbalance and prompt correlation characteristic of clinical cohorts. On the generative side, Diffusion Transformer (DiT) architectures offer a promising alternative to U-Net-based denoising, with stronger global coherence properties relevant to large abdominal field-of-view synthesis; additionally, spatial conditioning mechanisms such as ControlNet could provide structural grounding that partially compensates for the spatial ambiguity inherent in purely textual conditioning. Validating downstream utility --- whether synthetic volumes improve segmentation or classification under data augmentation --- and structured clinical evaluation by expert radiologists remain necessary steps before deployment.
 
\section{Limitations}
\label{sec:limitations}
OvESyn is evaluated as a synthesis framework for \emph{domain transfer}, with clinical deployment outside the current scope. The metrics we report --- FID2.5D, FID3D, Wasserstein-1 distance, mean intensity error, and Improved Precision and Recall --- quantify distributional realism, intensity fidelity, and the fidelity--coverage balance of the generated cohort; none of them certifies the oncologic plausibility of individual lesions. A volume can match the real distribution in feature space and intensity while still misrepresenting lesion morphology, margins, or the spatial relationships that carry diagnostic meaning. Establishing lesion-level validity is a separate undertaking that requires structured evaluation by expert radiologists --- including visual Turing tests and assessment of perceptual realism --- which we have not conducted.
 
Three further limitations bound the present results. First, the cohort comprises 493 patients from a single center (IEO Milan), with no external validation; generalization to other acquisition protocols, scanner vendors, or patient populations remains untested. Second, the downstream utility of the synthesized volumes --- whether they improve segmentation or classification under data augmentation --- has not been measured, and strong distributional and intensity metrics do not by themselves imply such benefit. Third, the conditioning text is derived from automatic OvSeg and TotalSegmentator segmentations, so segmentation errors propagate directly into the descriptors and cannot be corrected downstream. Together these define the steps required before OvESyn could move from distributional synthesis toward clinical use.
 
\section{Conclusion}
\label{sec:conclusion}
We presented OvESyn, the first systematic adaptation of a text-conditioned 3D CT generation framework to ovarian cancer, centered on a structured report generation pipeline that derives standardized conditioning text from CT-based imaging descriptors and routine preoperative metadata, without requiring original radiology reports.
 
On a cohort of 493 HGSOC patients, a controlled ablation identifies generator domain adaptation as the operative mechanism for crossing the domain gap: in its absence, synthesis remains anchored to the thoracic pretraining domain and fails to produce target-domain anatomy, irrespective of encoder alignment, whereas full generator fine-tuning yields anatomically coherent abdomino-pelvic volumes (FID2.5D below 33). Encoder alignment is complementary, refining intensity and per-sample fidelity within the established anatomy; the full configuration attains the best distributional realism and intensity fidelity (FID2.5D 29.35, Precision 0.671, Wasserstein-1 0.044) and is therefore the preferred model. Requiring only automatic segmentations and two routinely available clinical variables, OvESyn provides a reproducible and institutionally transferable basis for synthetic cohort generation in abdomino-pelvic oncologic imaging.
 
\section*{Acknowledgment}
This paper is supported by Fondazione Regionale per la Ricerca Biomedica (Regione Lombardia), project ID 012024R0055 PREDICT, and by the FAIR (Future Artificial Intelligence Research) project, funded by the NextGenerationEU program within the PNRR-PE-AI scheme (investment I.4.1). Model training and evaluation were enabled by the HPC resources provided through the e-INFRA CZ infrastructure (ID:90254).
 
\bibliographystyle{plainnat}
\bibliography{references}
 
\clearpage
 
\appendix
\section{Prompt Templates for Structured Report Generation}
\label{app:prompts}
Report generation is implemented as a two-stage one-shot prompting pipeline using Qwen2.5-14B-Instruct \cite{yang2025qwen3}, with separate templates for the Findings and Impression sections. Each template includes one in-context example and a set of hard constraints enforced at the prompt level. Runtime variables are populated from CT-derived features and clinical metadata prior to inference. The complete templates are reported below and can be used directly to replicate the conditioning pipeline.
 
\subsection*{A.1 Findings Template}
 
{\footnotesize
\begin{verbatim}
You are a radiologist writing objective CT findings.
Output ONLY the findings text within
<FINDINGS></FINDINGS> tags.
 
CRITICAL RULES - NO EXCEPTIONS:
1. NEVER use: "invades", "invasion", "metastatic",
   "malignancy", "peritoneal implants"
2. For organ contact, use ONLY: "is adjacent to [organ]" or 
"adjacent to [organ]" (do not imply invasion)
3. Ascites: [ASCITES_INSTRUCTION]
   (NEVER "mild/moderate/severe")
4. FIGO: State only "Findings are consistent with
   FIGO Stage [FIGO_STAGE] disease."
   (do not elaborate)
5. Use ONLY information from PATIENT DATA below -
   no additional clinical interpretation
6. Report individual volumes AND combined total
 
EXAMPLE:
<FINDINGS>
An irregular multilobulated omental mass measuring
14.9 cm with volume 31.2 mL demonstrates
predominantly solid attenuation with moderate
heterogeneity. The mass abuts small bowel and colon.
A separate lobulated pelvic/ovarian mass measuring
7.6 cm with volume 7.6 mL shows mixed solid and
cystic components with mild heterogeneity, abutting
bladder. Combined tumor burden is 38.8 mL. Ascites
is present. Findings are consistent with FIGO Stage
III disease.
</FINDINGS>
 
PATIENT DATA:
[TUMOR_BLOCK]
 
Combined tumor burden: [TOTAL_VOLUME_ML] mL
FIGO stage: [FIGO_STAGE]
Ascites: [ASCITES]
 
<FINDINGS>
\end{verbatim}
}
 
\noindent where \texttt{[TUMOR\_BLOCK]} is populated at runtime with one entry per lesion in the following format:
 
{\footnotesize
\begin{verbatim}
Tumor N ([LOCATION]):
- Size: [SIZE_DESC] measuring [SIZE_CM]
- Individual volume: [VOLUME_ML] mL
- Morphology: [SHAPE_DESC]
- Attenuation: [DENSITY_PRIMARY], [HETEROGENEITY]
- Mean HU: [MEAN_HU] (range [P10_HU]-[P90_HU])
- Low attenuation areas present  [if applicable]
- Adjacent organs: [ORGAN_LIST]
\end{verbatim}
}
 
\noindent and \texttt{[ASCITES\_INSTRUCTION]} resolves to \textit{``State: `Ascites is present.'\,''} or \textit{``State: `No ascites is identified.'\,''} depending on the clinical metadata value.
 
\subsection*{A.2 Impression Template}
{\footnotesize
\begin{verbatim}
You are writing the IMPRESSION section of a CT radiology
report from structured imaging descriptors for ovarian
cancer assessment.
 
Output ONLY within <IMPRESSION></IMPRESSION> tags.
 
TASK:
Generate a concise radiology-style IMPRESSION based only
on the KEY DATA provided.
 
RULES:
1. Summarize the clinically relevant interpretation;
   do not restate the findings line-by-line.
2. Write 1-2 sentences, 25-45 words total.
3. Mention overall distribution, total tumor burden,
   ascites status, and FIGO stage.
4. Do not mention individual lesion sizes, individual
   lesion volumes, HU values, organ-by-organ adjacency,
   or measurement-by-measurement details.
5. Use "adnexal/pelvic" instead of "pelvis/ovaries".
6. Do not add information not present in KEY DATA.
7. Do not diagnose histology, tissue invasion, metastases,
   or metastatic disease unless explicitly included in
   KEY DATA.
8. Avoid artificial phrases such as "imaging descriptors
   show", "clinical correlation is", "the key data indicate",
   and "findings are consistent with".
9. If ascites is absent, state "without ascites" or
   "no ascites"; if present, state "with ascites".
10. If attenuation or heterogeneity descriptors differ
    across regions, summarize the overall pattern rather
    than listing every component.
 
STYLE:
Use natural CT radiology language: concise, interpretive,
and non-repetitive.
 
EXAMPLE 1:
KEY DATA:
- Sites: omental and adnexal/pelvic
- Number of tumor regions: 2
- Total tumor burden: 38.8 mL
- Dominant region: adnexal/pelvic
- Overall distribution: multifocal
- Dominant attenuation pattern: mixed solid-cystic and
  predominantly solid
- Heterogeneity: moderate
- Ascites: present
- Interpretive pattern: advanced omental and
  adnexal/pelvic disease pattern with ascites
- FIGO stage: III
 
<IMPRESSION>
Multifocal omental and adnexal/pelvic tumor burden
totaling 38.8 mL, with ascites. Overall heterogeneous
solid/mixed pattern is in keeping with FIGO stage III
disease.
</IMPRESSION>
 
EXAMPLE 2:
KEY DATA:
- Sites: omental and adnexal/pelvic
- Number of tumor regions: 2
- Total tumor burden: 526.4 mL
- Dominant region: omental
- Overall distribution: extensive multifocal
- Dominant attenuation pattern: mixed solid-cystic
- Heterogeneity: marked
- Ascites: absent
- Interpretive pattern: advanced omental and
  adnexal/pelvic disease pattern
- FIGO stage: III
 
<IMPRESSION>
Extensive omental-dominant and adnexal/pelvic tumor
burden totaling 526.4 mL, without ascites. Overall
markedly heterogeneous mixed solid-cystic pattern is in
keeping with FIGO stage III disease.
</IMPRESSION>
 
EXAMPLE 3:
KEY DATA:
- Sites: adnexal/pelvic
- Number of tumor regions: 1
- Total tumor burden: 167.6 mL
- Dominant region: adnexal/pelvic
- Overall distribution: localized
- Dominant attenuation pattern: complex cystic
- Heterogeneity: mild
- Ascites: absent
- Interpretive pattern: localized adnexal/pelvic disease
  pattern
- FIGO stage: I
 
<IMPRESSION>
Localized adnexal/pelvic tumor burden totaling 167.6 mL,
without ascites. Complex cystic morphology with mild
heterogeneity is in keeping with FIGO stage I disease.
</IMPRESSION>
 
Now generate the IMPRESSION.
 
KEY DATA:
- Sites: [SITES]
- Number of tumor regions: [NUMBER_OF_TUMOR_REGIONS]
- Total tumor burden: [TOTAL_TUMOR_BURDEN] mL
- Dominant region: [DOMINANT_REGION]
- Overall distribution: [OVERALL_DISTRIBUTION]
- Dominant attenuation pattern: [DOMINANT_ATTENUATION_PATTERN]
- Heterogeneity: [HETEROGENEITY]
- Ascites: [ASCITES_STATUS]
- Interpretive pattern: [INTERPRETIVE_PATTERN]
- FIGO stage: [FIGO_STAGE]
 
<IMPRESSION>
\end{verbatim}
}
 
\noindent For the Impression stage, the prompt is conditioned on compact
structured key data rather than measurement-level free text. These key data
summarize the involved sites, number of tumor regions, total tumor burden,
dominant region, overall distribution, dominant attenuation pattern,
heterogeneity, ascites status, interpretive pattern, and FIGO stage.
 
\noindent The dominant region is defined as the site of the largest tumor
region by volume, while total tumor burden is computed as the sum of all
tumor-region volumes. Overall distribution is derived from the number of
tumor regions, distinct anatomical sites, and total burden, using descriptors
such as \textit{localized}, \textit{limited}, \textit{multifocal},
\textit{extensive localized}, and \textit{extensive multifocal}. Attenuation
and heterogeneity are summarized across regions to produce a concise
radiology-style Impression.
 
\noindent The generated Impression is constrained to one or two sentences and
25--45 words, and must mention disease distribution, total tumor burden,
ascites status, and FIGO stage. The prompt prevents unsupported additions
such as histologic diagnoses, tissue invasion, nodal involvement, metastases,
or organ-by-organ adjacency. Inference parameters were set to temperature
$0.3$, top-$p$ $0.9$, and a maximum of 400 tokens for Findings; and
temperature $0.2$, top-$p$ $0.9$, and a maximum of 120 tokens for Impression,
with repetition penalties of $1.1$ and $1.15$, respectively.
 
\section{Training Hyperparameters}
\label{app:hyperparams}
Table~\ref{tab:hyperparams} reports the full training configuration for all OvESyn components across all ablation configurations. Parameters shared across configurations are listed once.
 
\begin{table}[t]
\caption{Training hyperparameters for the fine-tuned OvESyn components.}
\label{tab:hyperparams}
\centering
\small
\begin{tabular}{lll}
\toprule
\textbf{Component} & \textbf{Parameter} & \textbf{Value} \\
\midrule
\textit{3D-CLIP} & Fine-tuning & LoRA \\
                 & Optimizer & AdamW \\
                 & Learning rate & $2\times10^{-5}$ \\
                 & Weight decay & 0.01 \\
                 & Scheduler & Cosine with warmup \\
                 & Warmup ratio & 0.05 \\
                 & Epochs & 100 \\
                 & GPUs & 8 \\
                 & Batch/rank & 3 \\
                 & Global batch & 24 \\
                 & Grad. accumulation & 1 \\
                 & Effective batch & 24 \\
                 & LoRA $r/\alpha/dropout$ & $16/32/0.2$ \\
                 & AMP & True \\
                 & Seed & 11 \\
\midrule
\textit{U-Net}   & Training & Full fine-tuning \\
                 & Optimizer & Adam \\
                 & Learning rate & $1\times10^{-4}$ \\
                 & Epochs & 200 \\
                 & GPUs & 8 \\
                 & Batch/rank & 3 \\
                 & Global batch & 24 \\
                 & CFG dropout & 0.1 \\
                 & Save frequency & 50 epochs \\
                 & Inference GPUs & 8 \\
                 & Inference batch/rank & 4 \\
                 & Inference steps & 30 \\
                 & Volume shape & $512\times512\times128$ \\
                 & Sampling scheduler & RFlow, $T=1000$ \\
                 & Guidance scale & 1.4 \\
                 & Sampling strategy & Uniform \\
\midrule
\textit{Hardware} & Precision & Mixed precision (AMP) \\
\bottomrule
\end{tabular}
\end{table}
 
\section{Extended Report Generation Examples}
\label{app:examples}
 
Tables~\ref{tab:report_example_stageIII_noascites_localized}--\ref{tab:report_example_stageIV_ascites_multifocal} show additional structured report generation examples covering distinct clinical configurations: a Stage~III patient without ascites and a single pelvic lesion (Table~\ref{tab:report_example_stageIII_noascites_localized}), a Stage~III patient with ascites and bilateral lesions (Table~\ref{tab:report_example_stageIII_ascites_multifocal}), and a Stage~IV patient with ascites and bilateral lesions (Table~\ref{tab:report_example_stageIV_ascites_multifocal}).
 
\begin{table*}[t]
\caption{Report generation example: Stage~III, ascites absent, localized mixed solid-cystic adnexal/pelvic disease.}
\label{tab:report_example_stageIII_noascites_localized}
\centering
\small
\begin{tabular}{p{0.13\textwidth} p{0.82\textwidth}}
\toprule
\textbf{Section} & \textbf{Generated Text} \\
\midrule
Findings & A large lobulated, elongated, flattened pelvic/ovarian mass measuring 11.3 cm with volume 151.8 mL demonstrates mixed solid and cystic components with moderate heterogeneity. This mass is adjacent to small bowel, colon, and urinary bladder. No ascites is identified. Findings are consistent with FIGO Stage III disease. \\
\midrule
Impression & Localized adnexal/pelvic tumor burden totaling 151.8 mL, without ascites. Mixed solid-cystic appearance with moderate heterogeneity suggests FIGO stage III disease. \\
\bottomrule
\end{tabular}
\end{table*}
 
\begin{table*}[t]
\caption{Report generation example: Stage~III, ascites present, multifocal predominantly solid disease.}
\label{tab:report_example_stageIII_ascites_multifocal}
\centering
\small
\begin{tabular}{p{0.13\textwidth} p{0.82\textwidth}}
\toprule
\textbf{Section} & \textbf{Generated Text} \\
\midrule
Findings & An irregular multilobulated omental mass measuring 22 cm with volume 55.1 mL demonstrates predominantly solid attenuation with moderate heterogeneity. The mass is adjacent to small bowel and colon. A separate irregular, multilobulated, elongated pelvic/ovarian mass measuring 11.0 cm with volume 49.6 mL shows predominantly solid attenuation with moderate heterogeneity. This mass is adjacent to small bowel, colon, and urinary bladder. Combined tumor burden is 104.8 mL. Ascites is present. Findings are consistent with FIGO Stage III disease. \\
\midrule
Impression & Multifocal omental and adnexal/pelvic tumor burden totaling 104.8 mL, with ascites. Predominantly solid pattern with moderate heterogeneity suggests FIGO stage III disease. \\
\bottomrule
\end{tabular}
\end{table*}
 
\begin{table*}[t]
\caption{Report generation example: Stage~IV, ascites present, multifocal predominantly solid disease.}
\label{tab:report_example_stageIV_ascites_multifocal}
\centering
\small
\begin{tabular}{p{0.13\textwidth} p{0.82\textwidth}}
\toprule
\textbf{Section} & \textbf{Generated Text} \\
\midrule
Findings & An irregular multilobulated omental mass measuring 24 cm with volume 137.4 mL demonstrates predominantly solid attenuation with moderate heterogeneity. The mass is adjacent to liver, stomach, small bowel, and colon. A separate lobulated pelvic/ovarian mass measuring 11.6 cm with volume 64.4 mL shows predominantly solid attenuation with moderate heterogeneity. This mass is adjacent to small bowel, colon, and urinary bladder. Combined tumor burden is 201.8 mL. Ascites is present. Findings are consistent with FIGO Stage IV disease. \\
\midrule
Impression & Multifocal omental and adnexal/pelvic tumor burden totaling 201.8 mL, with ascites. Predominantly solid pattern with moderate heterogeneity suggests FIGO stage IV disease. \\
\bottomrule
\end{tabular}
\end{table*}
 
\end{document}